\documentclass[runningheads]{llncs}
\usepackage{graphicx}
\usepackage{tikz}
\usepackage{comment}
\usepackage{amsmath,amssymb} 
\usepackage{color, colortbl}
\usepackage{wrapfig}

\usepackage[accsupp]{axessibility}  

\usepackage{algorithm}
\usepackage{algorithmic}

\usepackage{multirow}
\definecolor{mygray}{gray}{0.926}

\begin{document}
\pagestyle{headings}
\mainmatter
\def\ECCVSubNumber{3912} 

\title{Bi-level Feature Alignment for Versatile Image Translation and Manipulation}

\titlerunning{Bi-level Feature Alignment}

\author{Fangneng Zhan\inst{1,2}\thanks{denotes equal contribution, $^{\S}$ denotes corresponding author.}
\and
Yingchen Yu\inst{2}$^{\star}$ \and
Rongliang Wu\inst{2} \and
Jiahui Zhang\inst{2} \and
Kaiwen Cui\inst{2} \and
Aoran Xiao\inst{2} \and
Shijian Lu\inst{2}$^{\S}$  \and
Chunyan Miao\inst{2}
}

\authorrunning{F. Zhan et al.}
\institute{Nanyang Technological University, Singapore \\
\and
Max Planck Institute for Informatics, Germany\\
\email{fzhan@mpi-inf.mpg.de},
\email{\{shijian.lu,ascymiao\}@ntu.edu.sg} \\
\email{\{yingchen001,ronglian001,jiahui003,kaiwen001,aoran.xiao\}@e.ntu.edu.sg}
}

\maketitle

\begin{abstract}
Generative adversarial networks (GANs) have achieved great success in image translation and manipulation. However, high-fidelity image generation with faithful style control remains a grand challenge in computer vision. This paper presents a versatile image translation and manipulation framework that achieves accurate semantic and style guidance in image generation by explicitly building a correspondence. To handle the quadratic complexity incurred by building the dense correspondences, we introduce a bi-level feature alignment strategy that adopts a top-$k$ operation to rank block-wise features followed by dense attention between block features which reduces memory cost substantially. As the top-$k$ operation involves index swapping which precludes the gradient propagation, we approximate the non-differentiable top-$k$ operation with a regularized earth mover's problem so that its gradient can be effectively back-propagated. In addition, we design a novel semantic position encoding mechanism that builds up coordinate for each individual semantic region to preserve texture structures while building correspondences. Further, we design a novel confidence feature injection module which mitigates mismatch problem by fusing features adaptively according to the reliability of built correspondences. Extensive experiments show that our method achieves superior performance qualitatively and quantitatively as compared with the state-of-the-art.
\end{abstract}

\section{Introduction}\label{sec:introduction}
Image translation and manipulation aim to generate and edit photo-realistic images conditioning on certain inputs such as semantic segmentation \cite{park2019spade,wang2018pix2pixhd}, key points \cite{tang2019cycle,dong2018soft} and layout \cite{li2020bachgan}.
It has been studied intensively in recent years thanks to its wide spectrum of applications in various tasks~\cite{shrivastava2017learning,murez2018image,wan2020bringing}.
However, achieving high fidelity image translation and manipulation with faithful style control remains a grand challenge due to the high complexity of natural image styles.
A typical approach to control image styles is to encode image features into a latent space with certain regularization (e.g., Gaussian distribution) on the latent feature distribution.
For example, Park \emph{et al.}\cite{park2019spade} utilize VAE~\cite{doersch2016tutorial} to regularize the distribution of encoded features for faithful style control.
However, VAE struggles to encode the complex distribution of natural image styles and often suffers from \textit{posterior collapse}~\cite{lucas2019don} which leads to degraded style control performance. 
Another strategy is to encode reference images into style codes \cite{choi2020starganv2,zhu2020sean} to provide style guidance in image generation, while style codes often capture the global or regional style without an explicit style guidance for generating texture details.

\begin{figure}[t]
  \begin{minipage}[c]{0.475\textwidth}
    \includegraphics[width=\textwidth]{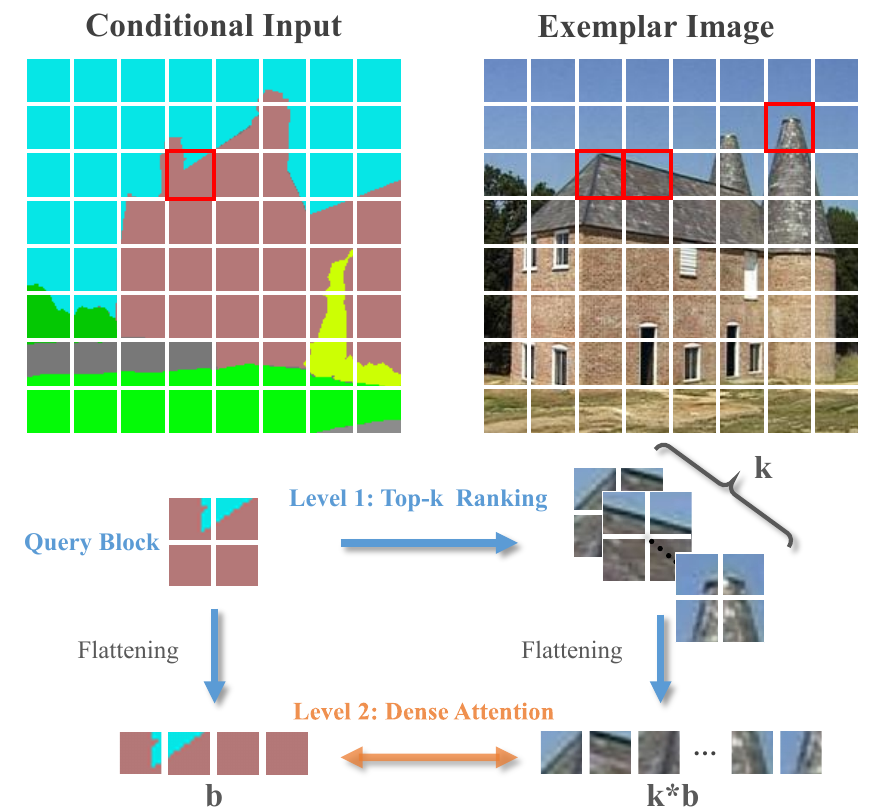}
  \end{minipage}\hfill
  \begin{minipage}[c]{0.45\textwidth}
    \caption{
    Bi-level feature alignment via ranking and attention scheme: With a query block from the \textit{Conditional Input}, we first retrieve the top-$k$ most similar blocks from the \textit{Exemplar Image} through a differentiable ranking operation, and then compute dense attention between features in query block and features in retrieved top-$k$ blocks. 
    Such bi-level alignment reduces the computational cost greatly, and allows to build high-resolution correspondences.
    } \label{im_intro}
  \end{minipage}
\end{figure}

To achieve more accurate style guidance and preserve details from exemplar, Zhang \emph{et al.} \cite{zhang2020cocosnet} explore to build cross-domain correspondences with Cosine similarity to achieve exemplar-based image translation. Zhou \emph{et al.} \cite{zhou2021cocosnetv2} propose a GRU-assisted Patch-Match \cite{barnes2009patchmatch} method to build high-resolution correspondences efficiently. Since the textures within a semantic region share identical semantic information, the existing methods tend to build correspondences based on the semantic coherence without considering the structure coherence within each semantic region. Warping exemplars with such pure semantic correspondence may cause destroyed texture patterns in the warped exemplars, and consequently result in inaccurate guidance for image generation. 

This paper presents \textbf{RABIT}, a \textbf{R}anking and \textbf{A}ttention scheme with \textbf{B}i-level feature alignment for versatile \textbf{I}mage \textbf{T}ranslation and manipulation. 
To mitigate the quadratic computational complexity issue of building the dense correspondence between conditional inputs (semantic guidance) and exemplars (style guidance), we design a bi-level alignment strategy with a Ranking and Attention Scheme (RAS) which builds feature correspondences efficiently at two levels: 1) a top-$k$ ranking operation for dynamically generating block-wise ranking matrices; 2) a dense attention module that achieves dense correspondences between features within blocks as illustrated in Fig. \ref{im_intro}.
RAS enables to build high-resolution correspondences and reduces the memory cost from $\mathcal{O}(L^2)$ to $\mathcal{O}(N^2 + b^2)$ ($L$ is the number of features for alignment, $b$ is block size, and $N=\frac{L}{b}$).
However, the top-$k$ operation involves index swapping whose gradient cannot be back-propagated in networks. To address this issue, we approximate the top-$k$ ranking operation with a regularized Earth Mover's problem \cite{emd} which enables gradient back-propagation effectively.

As in~\cite{zhang2020cocosnet,zhou2021cocosnetv2}, building correspondences based on semantic information only often leads to the losing of texture structures and patterns in warped exemplars.
Thus, spatial information should also be incorporated to preserve the texture structures and patterns and yield more accurate feature correspondences.
A vanilla method to encode the position information is concatenating the semantic features with the corresponding feature coordinates via coordconv \cite{liu2018coordconv}.
However, the vanilla position encoding builds a single coordinate system for the whole image which ignores the position information within each semantic region.
Instead, we design a semantic position encoding (SPE) mechanism that builds a dedicated coordinate system for each semantic region which outperforms the vanilla position encoding significantly.

In addition, conditional inputs and exemplars are seldom perfectly matched, e.g., conditional inputs could contain several semantic classes that do not exist in exemplar images.
Under such circumstances, the built correspondences often contain errors which lead to inaccurate exemplar warping and further deteriorated image generation.
We tackle this problem by designing a CONfidence Feature Injection (CONFI) module that fuses the features of conditional inputs and warped exemplars according to the reliability of the built correspondences.
Although the warped exemplar may not be reliable, the conditional input always provides accurate semantic guidance in image generation.
The CONFI module thus assigns higher weights to the conditional input when the built correspondence (or warped exemplar) is unreliable.
Experiments show that CONFI helps to generate faithful yet high-fidelity images consistently.

The contributions of this work can be summarized in three aspects.
First, we propose a versatile image translation and manipulation framework which introduces a ranking and attention Scheme for bi-level feature alignment that greatly reduces the memory cost while building the correspondence between conditional inputs and exemplars.
Second, we introduce a semantic position encoding 
mechanism that encodes region-level position information to preserve texture structures and patterns.
Third, we design a confidence feature injection 
module that provides reliable feature guidance in image translation and manipulation.

\section{Related Work} 

\subsection{Image-to-Image Translation}
Image translation has achieved remarkable progress in learning the mapping between images of different domains. It could be applied in different tasks such as style transfer~ \cite{huang2017adain,gatys2016image,li2017universal}, image super-resolution \cite{ledig2017photo,lim2017enhanced,lai2017deep,zhang2021blind}, domain adaptation \cite{shrivastava2017simgan,murez2018image,hoffman2018cycada,tsai2018learning,zhan2019gadan}, image composition \cite{zhan2019sfgan,zhan2020aicnet,zhan2021emlight,zhan2021gmlight,zhan2021sparse} etc. To achieve high-fidelity and flexible translation, existing work uses different conditional inputs such as semantic segmentation \cite{isola2017pix2pix,wang2018pix2pixhd,park2019spade,zhan2022marginal,zhan2022modulated}, scene layouts \cite{sun2019lostgan,zhao2019layout2im,li2020bachgan,zhan2021multimodal}, key points \cite{ma2017pose,men2020adgan,dong2018soft,zhan2021unite}, edge maps \cite{isola2017pix2pix,fu2019edit}, etc. However, effective style control remains a challenging task in image translation.

Style control has attracted increasing attention in image translation and generation. Earlier works such as~\cite{kingma2013vae} regularize the latent feature distribution to control the generation outcome. However, they struggle to capture the complex textures of natural images.
Style encoding has been studied to address this issue. For example, \cite{huang2018multimodal} and \cite{ma2018exemplar} transfer style codes from exemplars to source images via adaptive instance normalization (AdaIN) \cite{huang2017adain}. 
\cite{choi2020starganv2} employs a style encoder for style consistency between exemplars and translated images.
\cite{zhu2020sean} designs semantic region-adaptive normalization (SEAN) to control the style of each semantic region individually.
However, encoding style exemplars tends to capture the overall image style and ignores the texture details in local regions.
To achieve accurate style guidance for each local region, Zhang \emph{et al.} \cite{zhang2020cocosnet} build dense semantic correspondences between conditional inputs and exemplars with Cosine similarity to capture accurate exemplar details.
To mitigate the quadratic complexity issue and enable high-resolution correspondence building, 
Zhou \emph{et al.} \cite{zhou2021cocosnetv2} introduce the GRU-assisted Patch-Match to efficiently establish the high-resolution correspondence.

\subsection{Semantic Image Editing}
The arise of generative adversarial network (GANs) brings revolutionary advances to image editing \cite{zhu2016generative,hong2018learning,ntavelis2020sesame,choi2018stargan,pumarola2018ganimation,wu2020cascade,wu2020leed,xia2021tedigan}. As one of the most intuitive representation in image editing, semantic information has been extensively investigated in conditional image synthesis.
For example, 
Park \emph{et al.} \cite{park2019spade} introduce spatially-adaptive normalization (SPADE) to inject guided features in image generation.
MaskGAN \cite{lee2020maskgan} exploits a dual-editing consistency as auxiliary supervision for robust face image manipulation.
Instead of directly learning a label-to-pixel mapping, Hong \emph{et al.} \cite{hong2018learning} propose a semantic manipulation framework HIM that generates images guided by a predicted semantic layout.
Upon this work, Ntavelis \emph{et al.} \cite{ntavelis2020sesame} propose SESAME which requires only local semantic maps to achieve image manipulation.
However, the aforementioned methods either only learn a global feature without local focus (e.g., MaskGAN \cite{lee2020maskgan}) or ignore the features in the editing regions of the original image (e.g., HIM \cite{hong2018learning}, SESAME \cite{ntavelis2020sesame}).
To better utilize the fine features in the original image, Zheng \emph{et al.} \cite{zheng2020semantic} adapt exemplar-based image synthesis framework CoCosNet \cite{zhang2020cocosnet} for semantic image manipulation by building a high-resolution correspondence between the original image and the edited semantic map.

\begin{figure*}[t]
\centering
\includegraphics[width=1.0\linewidth]{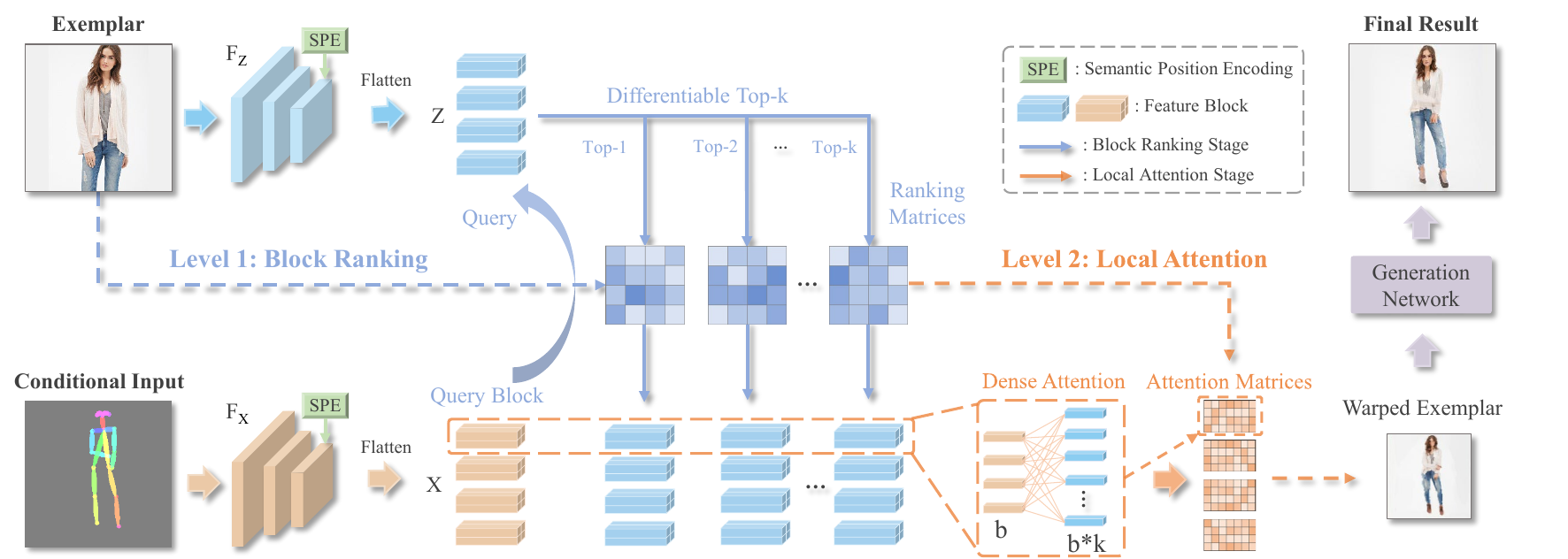}
\caption{
The framework of the proposed RABIT: \textit{Conditional Input} and \textit{Exemplar} are fed to feature extractors $F_{X}$ and $F_{Z}$ to extract feature vectors $X$ and $Z$ where $b$ local features form a feature block. 
In the first level, each block from the conditional input serves as the query to retrieve top-$k$ similar blocks from the exemplar through a differentiable ranking operation. 
In the second level, \textit{Dense Attention} is then built between the $b$ features in query block and $b*k$ features in the retrieved blocks. The built \textit{Ranking Matrices} and \textit{Attention Matrices} are combined to warp the exemplar to be aligned with the conditional input as in \textit{Warped Exemplar}, which serves as a style guidance to generate the final result through a generation network.
}
\label{im_stru}
\end{figure*}

\section{Proposed Method}

The proposed RABIT consists of an alignment network and a generation network that are inter-connected as shown in Fig. \ref{im_stru}. The alignment network learns the correspondence between a conditional input and an exemplar for warping the exemplar to be aligned with the conditional input. The generation network produces the final generation under the guidance of the warped exemplar and the conditional input.
RABIT is typically applicable in the task of conditional image translation with extra exemplar as style guidance.
It is also applicable to the task of image manipulation by treating the exemplars as the original images for editing and the conditional inputs as the edited semantic. The detailed loss functions can be found in the supplementary materials.
 
\subsection{Alignment Network}
The alignment network aims to build the correspondence between conditional inputs and exemplars, and accordingly provide accurate style guidance by warping the exemplars to be aligned with the conditional inputs.
As shown in Fig. \ref{im_stru}, conditional input and exemplar are fed to feature extractors $F_{X}$ and $F_{Z}$ to extract two sets of feature vectors $X = [x_1, \cdots, x_{L}] \in \mathbb{R}^{d}$ and $Z = [z_1,\cdots,z_{L}] \in \mathbb{R}^d$, where $L$ and $d$ denote the number and dimension of feature vectors, respectively.
Then $X$ and $Z$ can be aligned by building a $L\times L$ dense correspondence matrix where each entry denotes the Cosine similarity between the corresponding feature vectors in $X$ and $Z$.

\begin{wrapfigure}{r}{0.45\textwidth}
\centering
\includegraphics[width=1.0\linewidth]{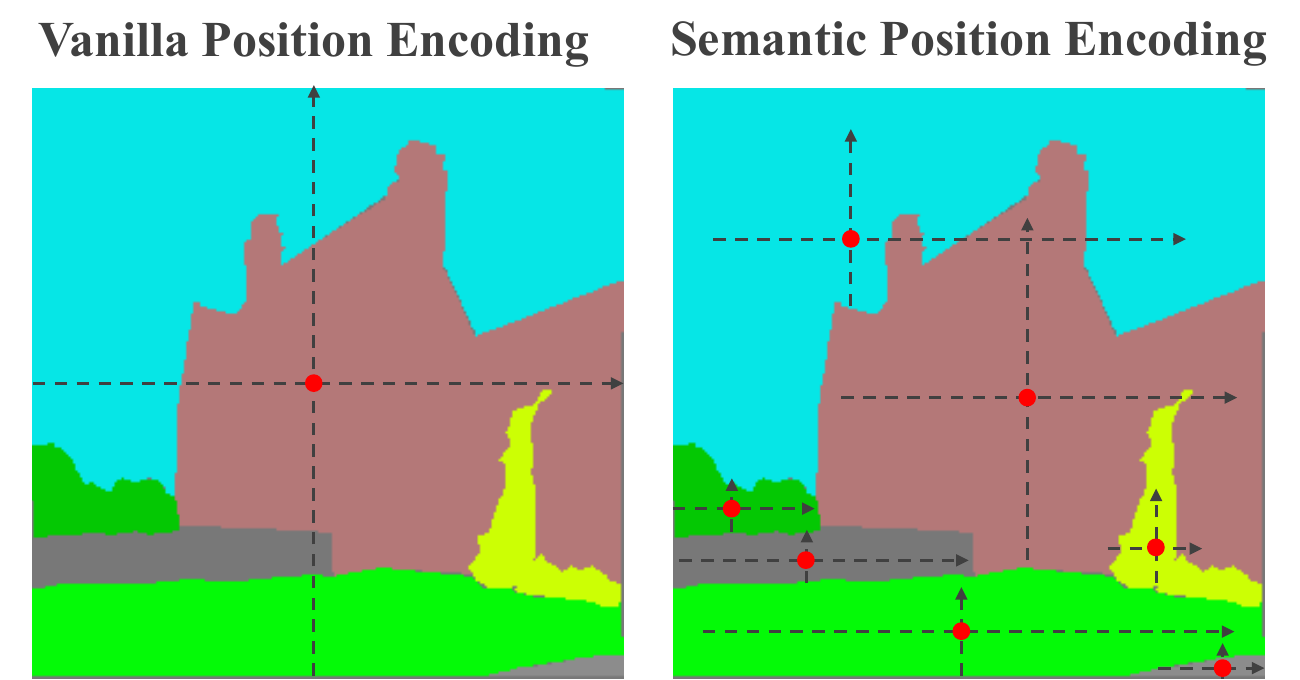}
\caption{
The comparison of vanilla position encoding and the proposed semantic position encoding (SPE). Red dots denote the coordinate origin. 
}
\label{im_spe}
\end{wrapfigure}

\begin{figure*}[t]
\centering
\includegraphics[width=1.0\linewidth]{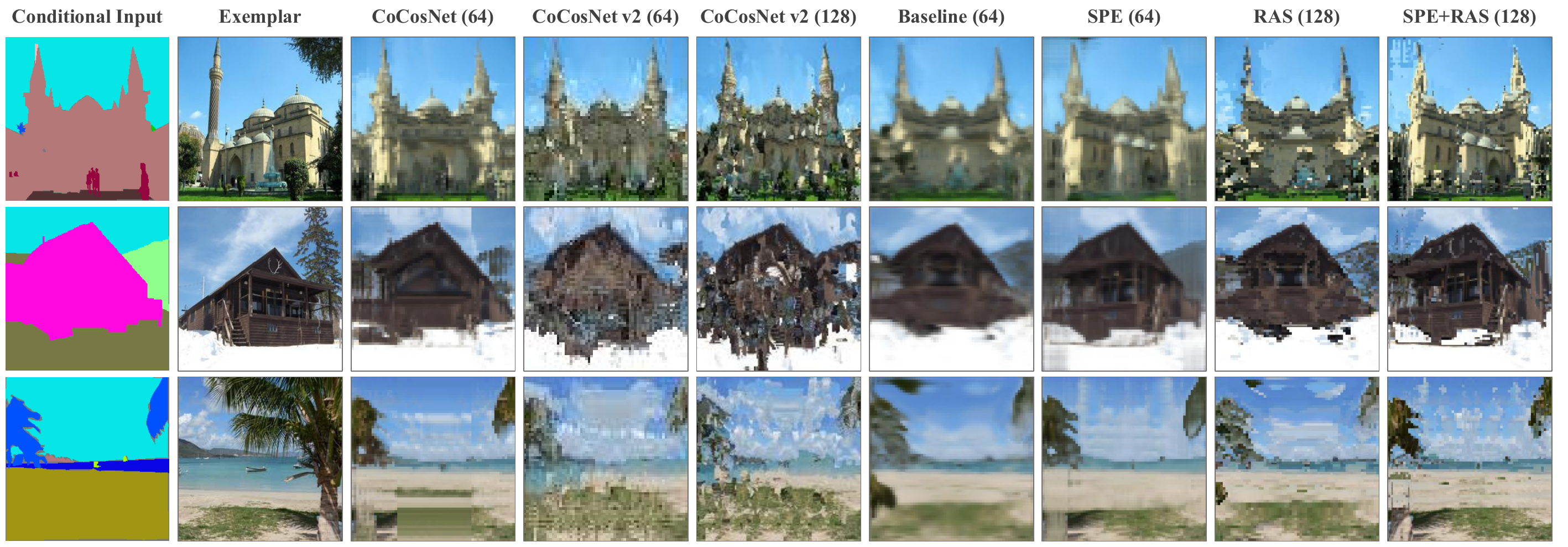}
\caption{
Warped exemplars with different methods: 
`64' and `128' mean to build correspondences at resolutions $64 \times 64$ and $128 \times 128$.
CoCosNet \cite{zhang2020cocosnet} tends to lose texture details and structures, while CoCosNet v2 \cite{zhou2021cocosnetv2} tends to generate messy warping.
The \textit{Baseline} denotes building correspondences with Cosine similarity, which tends to lose textures details and structures.
The proposed ranking and attention scheme (RAS) allows efficient image warping at high resolutions, the proposed semantic position encoding (SPE) can better preserve texture structures. The combination of the two as denoted by SPE+RAS achieves the best warping performance with high resolution and preserved texture structures. 
}
\label{im_warp}
\end{figure*}

\textbf{Semantic Position Encoding.}
Existing works \cite{zhang2020cocosnet,zhou2021cocosnetv2} mainly rely on semantic features to establish the correspondences. 
However, as textures within a semantic region share the same semantic feature, the pure semantic correspondence fails to preserve the texture structures or patterns within each semantic region.
Thus, the position information of features can be facilitated to preserve the texture structures and patterns. A vanilla method to encode the position information is employing a simple coordconv \cite{liu2018coordconv} to build a global coordinate for the full image.
However, this vanilla position encoding mechanism builds a single coordinate system for the whole image, ignoring region-wise semantic differences.
To preserve the fine texture pattern within each semantic region,
we design a semantic position encoding (SPE) mechanism that builds a dedicated coordinate for each semantic region as shown in Fig. \ref{im_spe}. 
Specifically, SPE treats the center of each semantic region as the origin of coordinate, and the coordinates within each semantic region are normalized to [-1, 1].
The proposed SPE outperforms the vanilla position encoding significantly as shown in Fig. \ref{im_warp} and to be evaluated in experiments.

\textbf{Bi-level Feature Alignment.}
On the other hand, building correspondence has quadratic complexity which incurs large memory and computation costs. 
Most existing studies thus work with low-resolution exemplar images (e.g. 64 $\times$ 64 in CoCosNet \cite{zhang2020cocosnet}) which often struggle in generating realistic images with fine texture details.
In this work, we propose a bi-level alignment strategy via a novel ranking and attention scheme (RAS) that greatly reduces computational costs and allows to build correspondences with high-resolution images as shown in Fig. \ref{im_warp}. 
Instead of building correspondences between features directly, the bi-level alignment strategy builds correspondences at two levels, including the first level that introduces top-$k$ ranking to generate block-wise ranking matrices dynamically and the second level that achieves dense attention between the features within blocks.
As Fig. \ref{im_stru} shows, $b$ local features are grouped into a block, thus the features of conditional input and exemplar are partitioned into $N$ blocks ($N = L/b$) as denoted by $X = [X_1, \cdots, X_{N}] \in \mathbb{R}^{bd}$ and $Z = [Z_1, \cdots, Z_{N}] \in \mathbb{R}^{bd}$.
In the first level of top-$k$ ranking, each block feature of the conditional input serves as a query to retrieve top-$k$ block features from the exemplar according to the Cosine similarity between blocks.
In the second level of local attention, the features in each query block further attends to the features in the top-$k$ retrieved blocks to build up local attention matrices within the block features.
The correspondence between the exemplar and conditional input can thus be built much more efficiently by combining such inter-block ranking and inner-block attention.

\begin{figure}[t!] 
\begin{minipage}[c]{0.48\linewidth}
\centering
\includegraphics[width=1.0\linewidth]{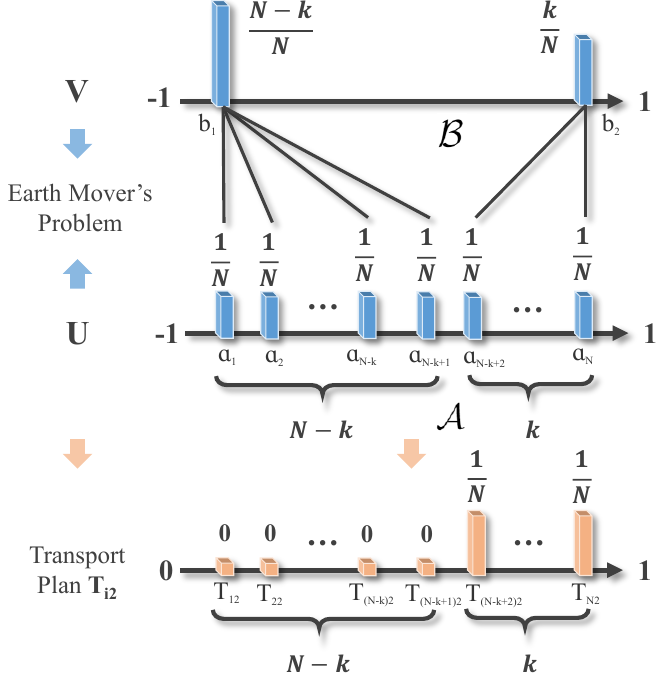}
\caption{
Illustration of the earth mover's problem in top-$k$ retrieval. Earth mover's problem is conducted between distributions $U$ and $V$ which is defined on supports $\mathcal{A}=[a_1,\cdots,a_N]$ and $\mathcal{B}=[b_1,b_2]$. \textit{Transport Plan $T_{i2}$} indicates the retrieved top-$k$ elements.
}
\label{im_ot}
\end{minipage}
\hfill
\begin{minipage}[c]{0.47\linewidth}
\centering
\includegraphics[width=1.0\linewidth]{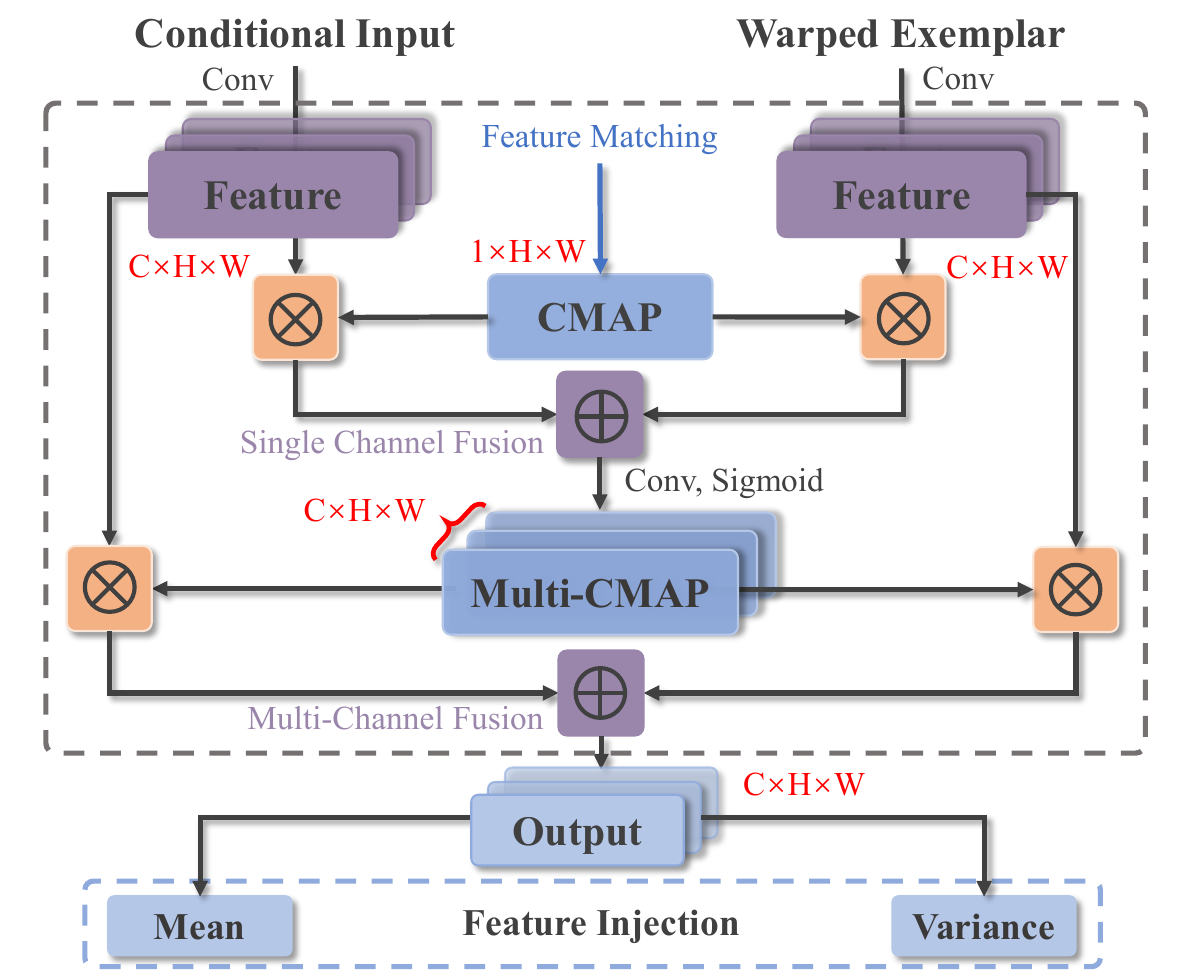}
\caption{
Illustration of confidence feature injection: 
Conditional input and warped exemplar are initially fused with a confidence map (CMAP) of size $1\times H \times W$. A multi-channel confidence map (Multi-CMAP) of size $C\times H \times W$ is then obtained from the initial fusion which further fuses the conditional input and warped exemplar in multiple channels.
}
\label{im_confi}
\end{minipage}%
\end{figure}

The ranking and attention scheme employs a top-$k$ operation that ranks the correlative blocks. However, the original top-$k$ operation involves index swapping whose gradient cannot be computed and so cannot be integrated into end-to-end network training. 
Inspired by Xie \emph{et al.} \cite{xie2020differentiable},
we tackle this issue by formulating the top-$k$ ranking as a regularized earth mover's problem which allows gradient computation via implicit differentiation.
Earth mover's problem aims to find a transport plan that minimizes the total cost to transform one distribution to another. Consider two discrete distributions $U = [\mu_1, \dots, \mu_N]^\top$ and $V = [\nu_1, \dots, \nu_M]^\top$ defined on supports $\mathcal{A}=[a_1, \cdots, a_N]$ and $\mathcal{B}=[b_1,\cdots,b_M]$, with probability (or amount of earth) $\mathbb{P} (a_i) = \mu_i$ and $\mathbb{P}(b_j) = \nu_j$.
We define $C \in \mathbb{R}^{N \times M}$ as the cost matrix where $C_{ij}$ denotes the cost of transportation between $a_i$ and $b_j$, and $T$ as a transport plan where $T_{ij}$ denotes the amount of earth transported between $\mu_{i}$ and $\nu_{j}$. An earth mover's (EM) problem can be formulated by:
$ 
{\rm EM} = \mathop{\min}\limits_{T}  \langle C, T \rangle, \quad {\rm s.t.} \ T  \vec{1}_M = U, \; T^\top  \vec{1}_N = V
$
where $\vec{1}$ denotes a vector of ones, $\langle \rangle$ denotes inner product.
By treating a correlation scores between a query block and $N$ key blocks as $\mathcal{A}=[a_1, \cdots, a_N], a_{i} \in [-1,1]$ and defining $\mathcal{B}=\{-1, 1 \}$, $U=[\mu_1, \cdots, \mu_N]$ and $V = [\nu_1, \nu_2] $, it can be proved that solving the Earth Mover's problem is equivalent to select the largest $K$ elements from $\mathcal{A}=[a_1,\cdots,a_N]$. 
The detailed proof and optimization of the earth mover's problem is provided in supplementary material.
Fig. \ref{im_ot} illustrates the earth mover's problem and transport plan $T$ which indicates the top-$k$ elements.

\textbf{Complexity Analysis. }
The vanilla dense correspondence has a self-attention memory complexity $\mathcal{O}(L^2)$ where $L$ is the input sequence length. For our bi-level alignment strategy, the memory complexity of building block ranking matrices and local attention matrices are $\mathcal{O}(N^2)$ and $\mathcal{O}(b*(kb))$, where $b$, $N$ ($N=L/b$) and $k$ are block size, block number and the number of top-$k$ selection.
Thus, the overall memory complexity is $\mathcal{O}(N^2+b*(kb))$.

\subsection{Generation Network}

The generation network aims to synthesize images under the semantic guidance of conditional inputs and style guidance of exemplars.
The overall architecture of the generation network is similar to SPADE \cite{park2019spade}. Please refer to supplementary material for details of the network structure.

State-of-the-art approach~\cite{zhang2020cocosnet} simply concatenates the warped exemplar and conditional input to guide the image generation process. 
However, the warped input image and edited semantic map could be structurally aligned but semantically different especially when they have severe semantic discrepancy. Such unreliably warped exemplars could serve as false guidance and heavily deteriorate the generation performance. 
Therefore, a mechanism is required to identify the semantic reliability of warped exemplar to provide reliable guidance for the generation network.
To this end, we propose a CONfidence Feature Injection (CONFI) module that adaptively weights the features of conditional input and warped exemplar according to the reliability of feature matching.

\textbf{Confidence Feature Injection.} 
Intuitively, in the case of lower reliability of the feature correspondence, we should assign a relatively lower weight to the warped exemplar which provides unreliable style guidance and a higher weight to the conditional input which consistently provides accurate semantic guidance.

As illustrated in Fig. \ref{im_confi}, the proposed CONFI fuses the features of the conditional input and warped exemplar based on a confidence map (CMAP) that captures the reliability of the feature correspondence.
To derive the confidence map, we first obtain a block-wise correlation map of size $N \times N$ by computing element-wise Cosine distance between $X = [X_i, \cdots, X_N]$ and $Z = [Z_i, \cdots, Z_N]$. For a block $X_i$, the correlation score with $Z$ is denoted by $\mathcal{A}=[a_1, \cdots, a_N]$.
As higher correlation scores indicate more reliable feature matching, we treat the peak value of $\mathcal{A}$ as the confidence score of $X_i$.
Similar for other blocks, we can obtain the confidence map (CMAP) of size $1 \times H \times W$ ($N=H*W$) which captures the semantic reliability of all blocks.
The features of the conditional input and exemplar (both of size $C \times H \times W$ after passing through convolution layers) can thus be fused via weighted sum based on the confidence map CMAP:
$F = X * (1- {\rm CMAP}) + (T \cdot Z) * {\rm CMAP}$
where $T$ is the built correspondence matrix.
As the confidence map contains only one channel ($1 \times H  \times W$), the above feature fusion is conducted in $H\times W$ but ignores that in $C$ channel.
To achieve thorough feature fusion in all channels, we feed the initial fusion $F$ to convolution layers to generate a multi-channel confidence map (Multi-CMAP) of size $C \times H \times W$.
The conditional input and warped exemplar are then thoroughly fused via a full channel-weighted summation according to the Multi-CMAP.
The final fused feature is further injected to the generation process via spatial de-normalization \cite{park2019spade} to provide accurate semantic and style guidance.

\section{Loss Functions}
\label{loss}

The alignment network and generation network are jointly optimized.
For clarity, we still denote the conditional input and exemplar as $X$ and $Z$, the ground truth as $X'$, the generated image as $Y$,
the feature extractors for conditional input and exemplar as $E_{X}$ and $E_{Z}$, the generator and discriminator in the generation network as $G$ and $D$.

\textbf{Alignment Network.}
First, the warping should be cycle consistent, i.e. the exemplar should be recoverable from the warped warped. We thus employ a cycle-consistency loss as follows:
\begin{align*}
    \mathcal{L}_{cyc} = || T^{\top} \cdot T\cdot Z - Z ||_{1}
\end{align*}
where $T$ is the correspondence matrix.
The feature extractors $F_{X}$ and $F_{Z}$ aim to extract invariant semantic information across domains, i.e. the extracted features from $X$ and $X'$ should be consistent. 
A feature consistency loss can thus be formulated as follows:
\begin{align*}
    \mathcal{L}_{cst}= || F_{X}(X) - F_{Z}(X') ||_{1}
\end{align*}

\textbf{Generation Network.}
The generation network employs several losses for high-fidelity synthesis with consistent style with the exemplar and consistent semantic with the conditional input.
As the generated image $Y$ should be semantically consistent with the ground truth $X'$, we employ a perceptual loss $\mathcal{L}_{perc}$ \cite{johnson2016perceptual} to penalize their semantic discrepancy as below: 
\begin{equation}
    \mathcal{L}_{perc} = || \phi_{l}(Y) - \phi(X') ||_{1}
\end{equation}
where $\phi_{l}$ is the activation of layer $l$ in pre-trained VGG-19 \cite{simonyan2014vgg} model. To ensure the statistical consistency between the generated image $Y$ and the exemplar $Z$, a contextual loss \cite{mechrez2018contextual} is adopted:
\begin{equation}
    \mathcal{L}_{cxt} = - \log( \sum_{i} \mathop{\max}\limits_{j} CX_{ij} ( \phi_{l}^{i} (Z), \phi_{l}^{j} (Y) ) ) 
\end{equation}
where $i$ and $j$ are the indexes of the feature map in layer $\phi_{l}$.
Besides, a pseudo pairs loss $\mathcal{L}_{pse}$ as described in \cite{zhang2020cocosnet} is included in training.

The discriminator $D$ is employed to drive adversarial generation with an adversarial loss $\mathcal{L}_{adv}$ \cite{isola2017pix2pix}.
The full network is thus optimized with the following objective:
\begin{equation}
\begin{split}
    \mathcal{L} = & \mathop{\min}\limits_{F_{X},F_{Z},G} \mathop{\max}\limits_{D} (\lambda_1 \mathcal{L}_{cyc} + \lambda_2 \mathcal{L}_{cst} + \lambda_3 \mathcal{L}_{perc} \\
    &  + \lambda_4 \mathcal{L}_{cxt} + \lambda_5  \mathcal{L}_{pse} + \lambda_6 \mathcal{L}_{adv}) \\
\end{split}
\end{equation}
where the weights $\lambda$ balance the losses in the objective.

\section{Experiments}

\subsection{Experimental Settings}
\label{setting}

\textbf{Datasets.}
We evaluate and benchmark our method over multiple datasets for image translation \& manipulation tasks.

$\bullet$ ADE20K \cite{zhou2017ade20k} is adopted for image translation conditioned on semantic segmentation.
For image manipulation, we apply object-level affine transformations on the test set to acquire paired data (150 images) for evaluations as in \cite{zheng2020semantic}.

$\bullet$ CelebA-HQ \cite{liu2015celebahq} 
is used for two translation tasks by using face semantics and face edges as conditional inputs. 
We use 2993 face images for translation evaluations as in \cite{zhang2020cocosnet}, and manually edit 100 semantic maps which is randomly selected for image manipulation evaluations.

$\bullet$ DeepFashion \cite{liu2016deepfashion} is used for image translation conditioned key points.

\noindent
\textbf{Implementation Details:}
The default size for our correspondence computation is $128 \times 128$ with a block size of $2 \times 2$. The number $k$ in top-$k$ ranking is set at 3 by default in our experiments. The default size of generated images is $256 \times 256$.

\subsection{Image Translation Experiments} 
We compare RABIT with several state-of-the-art image translation methods.

\renewcommand\arraystretch{1.2}
\begin{table*}[t]
\small 
\caption{
Comparing RABIT with state-of-the-art image translation methods over four translation tasks with FID, SWD and LPIPS as the evaluation metrics.
}
\renewcommand\tabcolsep{3pt}
\centering 
\resizebox{1.0\textwidth}{!}{
\begin{tabular}{l|ccc|ccc|ccc|ccc} 
\hline
\rowcolor{mygray} 
& 
\multicolumn{3}{c|}{\textbf{ADE20K}} & 
\multicolumn{3}{c}{\textbf{CelebA-HQ (Semantic)}} &
\multicolumn{3}{c|}{\textbf{DeepFashion}} &
\multicolumn{3}{c}{\textbf{CelebA-HQ (Edge)}}
\\
\cline{2-13}
\rowcolor{mygray} 
\multirow{-2}{*}{\textbf{Methods}} & FID $\downarrow$ & SWD $\downarrow$ & LPIPS $\uparrow$ & FID $\downarrow$ & SWD  $\downarrow$ & LPIPS  $\uparrow$ & FID  $\downarrow$ & SWD  $\downarrow$ & LPIPS  $\uparrow$ & FID  $\downarrow$ & SWD  $\downarrow$ & LPIPS  $\uparrow$  
\\\hline 

\textbf{Pix2pixHD\cite{wang2018pix2pixhd}} & 81.80 & 35.70 & N/A     & 43.69 & 34.82 & N/A      & 25.20  & 16.40  & N/A       & 42.70 & 33.30 & N/A         \\

\textbf{StarGAN v2\cite{choi2020starganv2}} & 98.72 & 65.47 & 0.551      &  53.20  & 41.87 &  \textbf{0.324}       & 43.29  & 30.87  & \textbf{0.296}     & 48.63 & 41.96 & \textbf{0.214}        \\

\textbf{SPADE\cite{park2019spade}} & 33.90 & 19.70 & 0.344       & 39.17 & 29.78 & 0.254      & 36.20  & 27.80  & 0.231  & 31.50 & 26.90 & 0.207        \\

\textbf{SelectionGAN\cite{tang2019selectiongan}}     &  35.10 & 21.82 & 0.382      & 42.41 & 30.32 & 0.277       & 38.31  & 28.21  & 0.223   & 34.67 & 27.34 &  0.191       \\

\textbf{SMIS\cite{zhu2020smis}} & 42.17 &  22.67 &  0.476       & 28.21 & 24.65 & 0.301        & 22.23  & 23.73  & 0.240       &  23.71  &  22.23  & 0.201      \\

\textbf{SEAN\cite{zhu2020sean}}     & 24.84 &  10.42  & 0.499       & \textbf{17.66} & 14.13 & 0.285       & 16.28  &  17.52  & 0.251       &  16.84    & 14.94 & 0.203        \\

\textbf{CoCosNet\cite{zhang2020cocosnet}}     & 26.40 & 10.50 & \textbf{0.580}        & 21.83 & 12.13  &  0.292       & 14.40  & 17.20 & 0.272        & 14.30 & 15.30 & 0.208         \\

\hline
\textbf{RABIT}
& \textbf{24.35} & \textbf{9.893} & \textbf{0.571}
& 20.44 & \textbf{11.18} & 0.307
& \textbf{12.58} & \textbf{16.03} & 0.284
& \textbf{11.67} & 14.22 & 0.209
  \\\hline
\end{tabular}}
\label{tab_com}
\end{table*}

\begin{figure*}[t]
\centering
\includegraphics[width=1.0\linewidth]{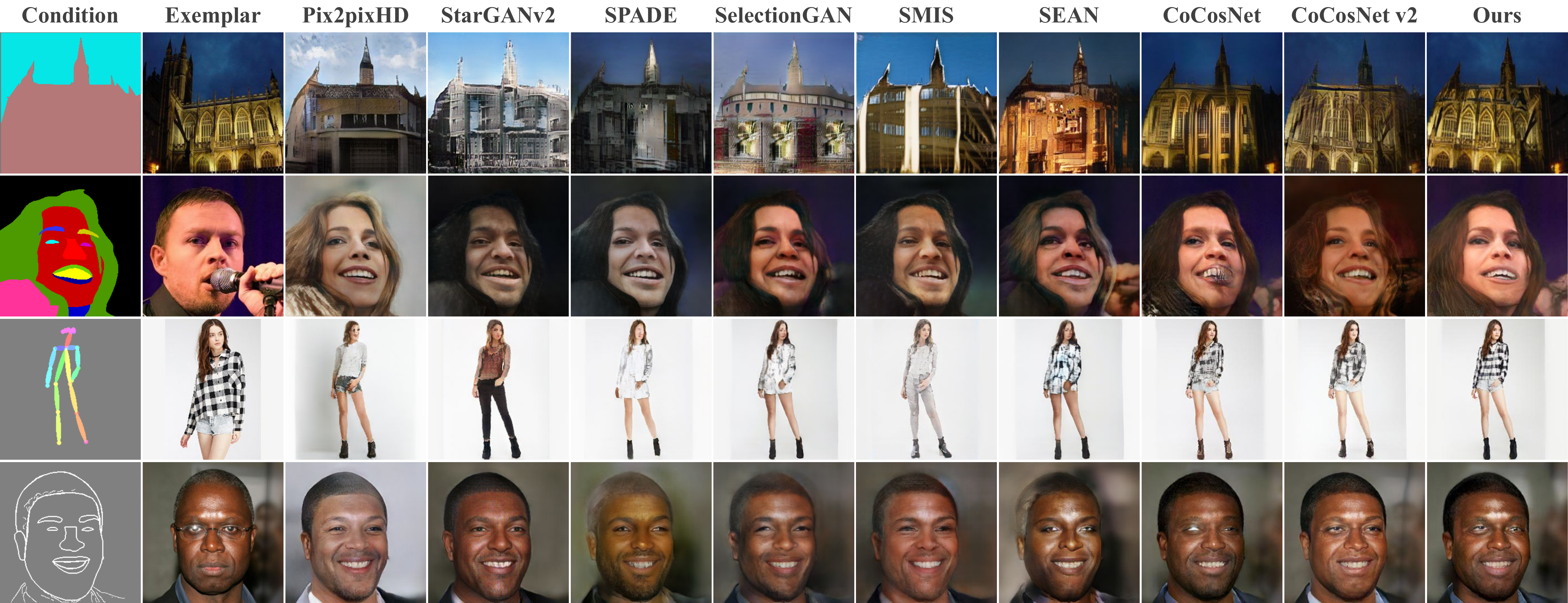}
\caption{
Qualitative comparison of the proposed RABIT and state-of-the-art methods over four types of conditional image translation tasks.
}
\label{im_tran_com}
\end{figure*}

\begin{figure*}[t]
\centering
\includegraphics[width=1.0\linewidth]{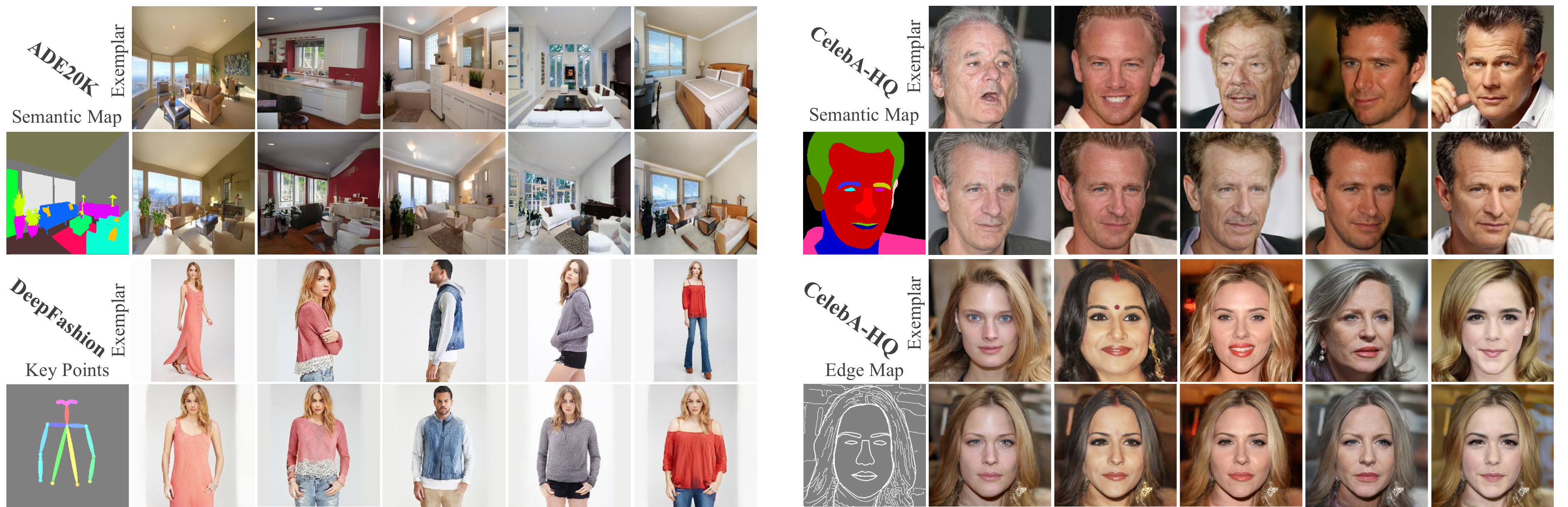}
\caption{
Illustration of generation diversity of RABIT: With the same conditional input, RABIT can generate a variety of images that have consistent styles with the provided exemplars. It works for different types of conditional inputs consistently.
}
\label{im_diverse}
\end{figure*}

\textbf{Quantitative Results.}
In quantitative experiments, all methods translate images with the same exemplars except Pix2PixHD \cite{wang2018pix2pixhd} which doesn't support style injection from exemplars. 
LPIPS is calculated by comparing the generated images with randomly selected exemplars.
All compared methods adopt three exemplars for each conditional input and the final LPIPS is obtained by averaging the LPIPS between any two generated images.

Table \ref{tab_com} shows experimental results. It can be seen that RABIT outperforms all compared methods over most metrics and tasks consistently.
By building explicit yet accurate correspondences between conditional inputs and exemplars, RABIT enables direct and accurate guidance from the exemplar and achieves better translation quality (in FID and SWD) and diversity (in LPIPS) as compared with the regularization-based methods such as SPADE \cite{park2019spade} and SMIS \cite{zhu2020smis}, and style-encoding methods such as StarGAN v2 \cite{choi2020starganv2} and SEAN \cite{zhu2020sean}. 
Compared with correspondence-based method CoCosNet \cite{zhang2020cocosnet}, the proposed bi-level alignment allows RABIT to build correspondences and warp exemplars at higher resolutions (e.g. $128 \times 128$) which offers more detailed guidance in the generation process and helps to achieve better FID and SWD.
While compared with CoCosNet v2 \cite{zhou2021cocosnetv2}, the proposed semantic position encoding enables to preserve the texture structures and patterns, thus yielding more accurate warped exemplars as guidance.
Besides generation quality, RABIT achieves the best generation diversity in LPIPS except StarGAN v2 \cite{choi2020starganv2} which sacrifices the generation quality with much lower FID and SWD.

\renewcommand\arraystretch{1.05}
\begin{table*}[t!]
\centering

\caption{
Comparing RABIT with state-of-the art image manipulation methods on ADE20K \cite{zhou2017ade20k} and CelebA-HQ \cite{liu2015celebahq}.
}
\label{tab_com_ade}
\renewcommand\tabcolsep{3pt}
\resizebox{1.0\textwidth}{!}{
\begin{tabular}{l | ccc || l | ccc} 
\hline
\rowcolor{mygray} 
\multicolumn{4}{c||}{\textbf{ADE20K \cite{zhou2017ade20k}}} & 
\multicolumn{4}{c}{\textbf{CelebA-HQ \cite{liu2015celebahq}}}  \\
\hline
\rowcolor{mygray} 
\textbf{Models} & \textbf{FID}  $\downarrow$ & \textbf{PSNR}  $\uparrow$ & \textbf{SSIM}  $\uparrow$ & \textbf{Models} & \textbf{FID}  $\downarrow$ & \textbf{SWD}  $\downarrow$ & \textbf{LPIPS}  $\downarrow$
\\
\cline{2-3}
\hline
\textbf{SPADE} \cite{park2019spade}     & 120.2   &  13.11   &  0.334 &\textbf{SPADE} \cite{park2019spade}     & 105.1  &  41.90  & 0.376\\
\textbf{HIM} \cite{hong2018learning}     & 59.89    & 18.23  & 0.667 & \textbf{SEAN} \cite{zhu2020sean}     & 96.31  & 35.90  & 0.351 \\
\textbf{SESAME} \cite{ntavelis2020sesame}   & 52.51    &  18.67  & 0.691 & \textbf{MaskGAN} \cite{lee2020maskgan}      & 80.89  &  23.86  & 0.271 \\
\textbf{CoCosNet} \cite{zhang2020cocosnet}  & 41.03   & 20.30 & 0.744 &\textbf{CoCosNet} \cite{zhang2020cocosnet}     & 68.70  & 22.90 &  0.224 \\
\hline
\textbf{RABIT}   & \textbf{26.61} &  \textbf{23.08} & \textbf{0.823} & \textbf{RABIT}   & \textbf{60.87} & \textbf{21.07} &  \textbf{0.176}\\
\hline
\end{tabular}}
\end{table*}

\begin{figure*}[t]
\centering
\includegraphics[width=1.0\linewidth]{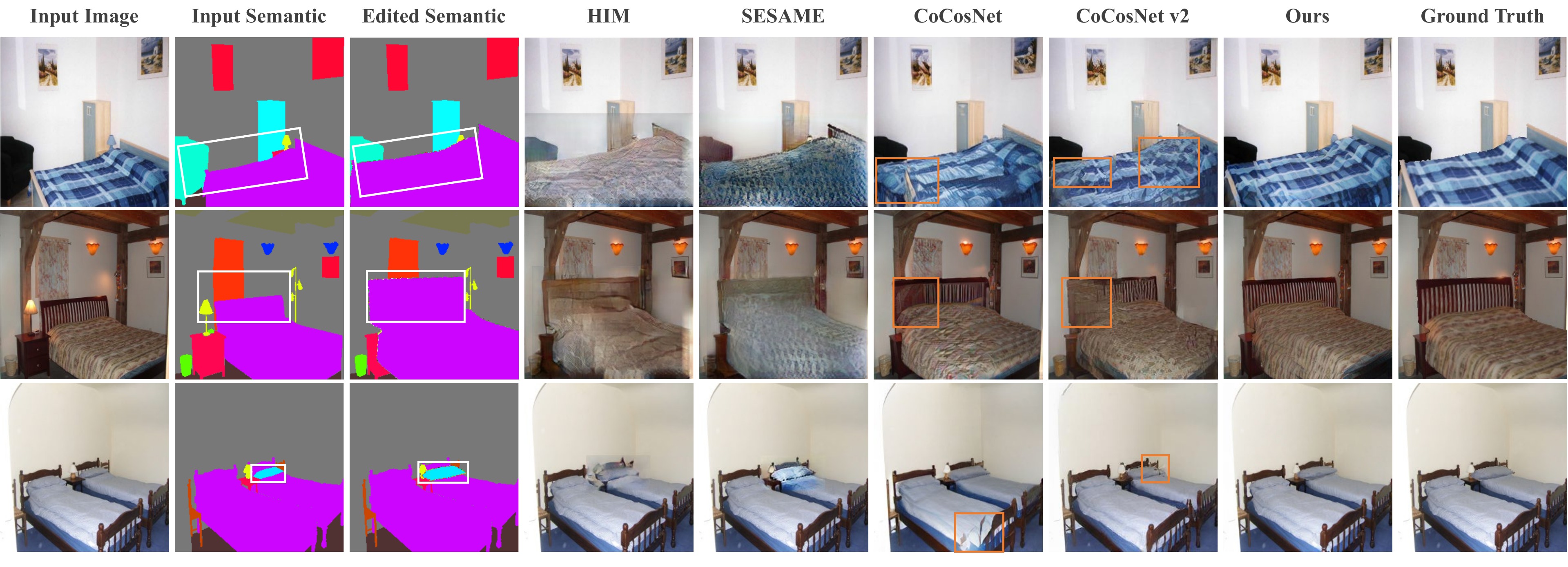}
\caption{
Qualitative illustration of RABIT and SOTA image manipulation methods on the augmented test set of ADE20K with ground truth as described in \cite{zheng2020semantic}.
}
\label{im_edit_ade}
\end{figure*}

\textbf{Qualitative Evaluations.}
Fig. \ref{im_tran_com} shows qualitative comparisons on various conditional image translation tasks. 
It can be seen that RABIT achieves the best visual quality with faithful styles as exemplars. 
RABIT also demonstrates superior diversity in image translation as illustrated in Fig. \ref{im_diverse}.

\subsection{Image Manipulation Experiment}

RABIT manipulates images by treating input images as exemplars and edited semantic guidance as conditional inputs. We compare RABIT with several state-of-the-art image manipulation methods 
including
1) SPADE \cite{park2019spade},
2) SEAN \cite{zhu2020sean},
3) MaskGAN \cite{CelebAMask-HQ},
4) Hierarchical Image Manipulation (HIM) \cite{hong2018learning}, 
5) SESAME \cite{ntavelis2020sesame}, 
6) CoCosNet \cite{zhang2020cocosnet}.

\begin{figure*}[t]
\centering
\includegraphics[width=1.0\linewidth]{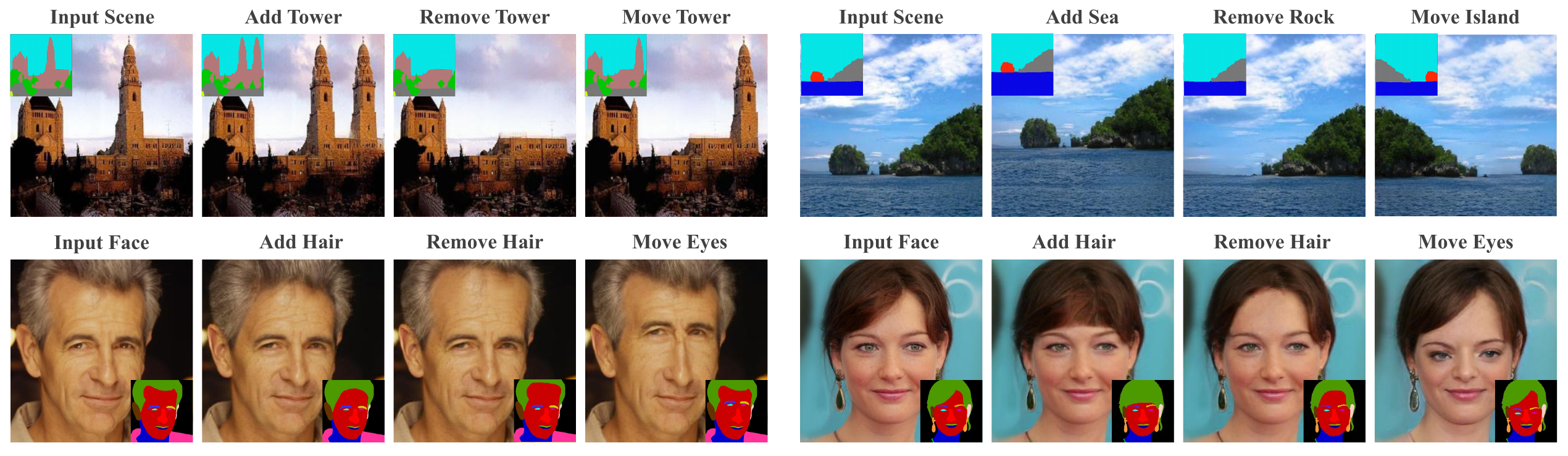}
\caption{
Various image editing by the proposed RABIT: With input images as the exemplars and edited semantic maps as the conditional input, RABIT generates new images with faithful semantics and high-fidelity textures with little artifacts.
}
\label{im_edit_diverse}
\end{figure*}

\textbf{Quantitative Results:}
In quantitative experiments, all compared methods manipulate images with the same input image and edited semantic label map.
Left side of Table \ref{tab_com_ade} shows experimental results over the synthesized test set of ADE20K \cite{zhou2017ade20k}. 
It can be observed that RABIT outperforms state-of-the-art methods over all evaluation metrics consistently.  
Right side of Table \ref{tab_com_ade} shows experimental results over the CelebA-HQ dataset with manually edited semantic maps. It can be observed that RABIT outperforms the state-of-the-art methods by large margins in all perceptual quality metrics.

\begin{wrapfigure}{r}{0.5\textwidth}
\centering
\includegraphics[width=1.0\linewidth]{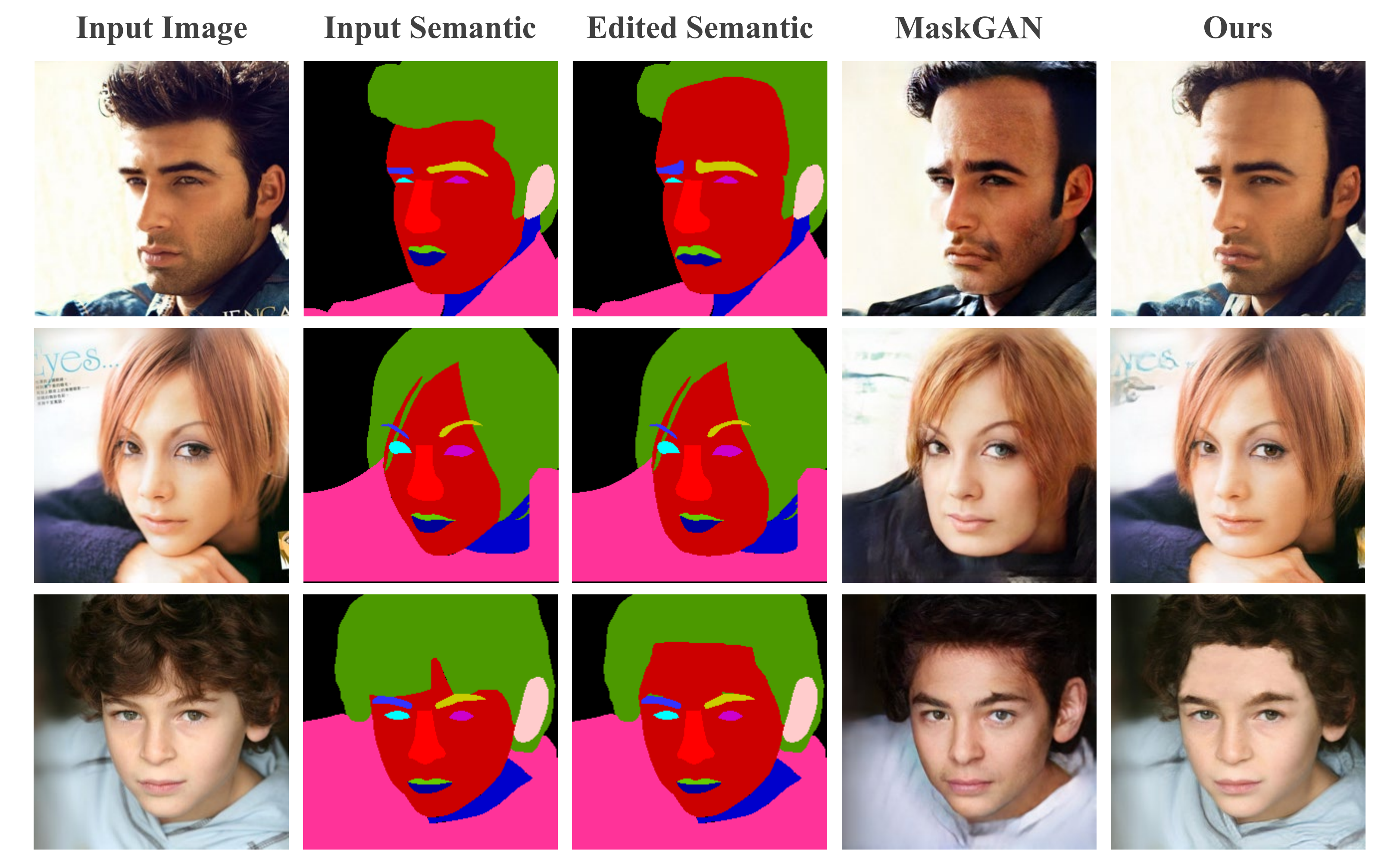}
\caption{
The comparison of image manipulation by MaskGAN \cite{lee2020maskgan} and the proposed RABIT over dataset CelebA-HQ \cite{liu2015celebahq}.
}
\label{im_edit_celebahq}
\end{wrapfigure}

\textbf{Qualitative Evaluation:}
Fig. \ref{im_edit_ade} shows visual comparisons with state-of-art manipulation methods on ADE20K. 
Fig. \ref{im_edit_diverse} shows the editing capacity of RABIT with various types of manipulation on semantic labels. 
We also compare RABIT with MaskGAN \cite{lee2020maskgan} on CelebA-HQ \cite{CelebAMask-HQ} in Fig. \ref{im_edit_celebahq}.

\begin{figure*}[ht]
\centering
\includegraphics[width=1.0\linewidth]{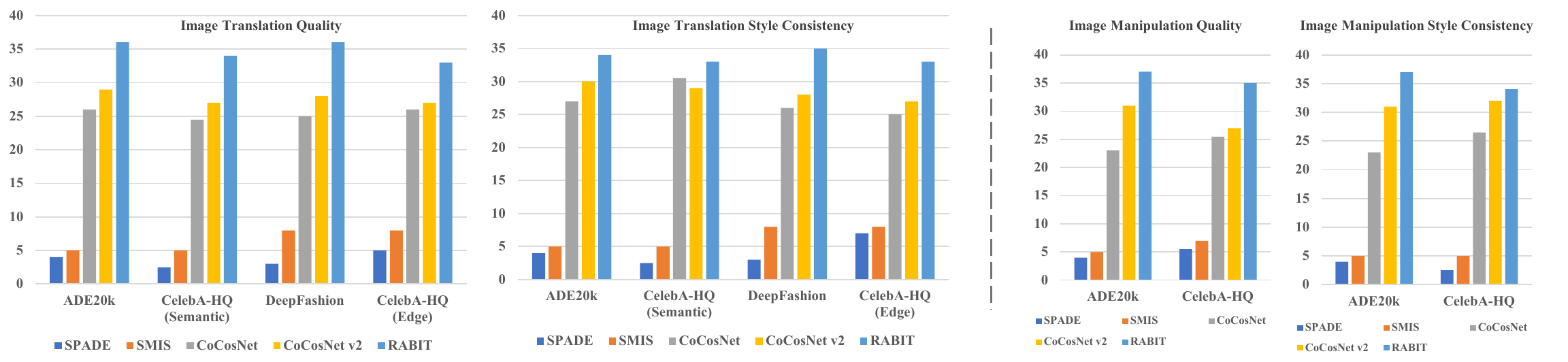}
\caption{
AMT (Amazon Mechanical Turk) user studies of different image translation and image manipulation methods in terms of the visual quality and style consistency of the generated images.
}
\label{im_amt}
\end{figure*}

\section{User Study}
\label{user}

We conduct crowdsourcing user studies through Amazon Mechanical Turk (AMT) to evaluate the image translation \& manipulation in terms of generation quality and style consistency.
Specifically, each compared method generates 100 images with the same conditional inputs and exemplars. Then the generated images together with the conditional inputs and exemplars were presented to 10 users for assessment. 
For the evaluation of image quality, the users were instructed to pick the best-quality images.
For the evaluation of style consistency, the users were instructed to select the images with best style relevance to the exemplar.
The final AMT score is the averaged number of the methods to be selected as the best quality and the best style relevance.

Fig.~\ref{im_amt} shows AMT results on multiple datasets.
It can be observed that RABIT outperforms state-of-the-art methods consistently in image quality and style consistency on both image translation \& image manipulation tasks.

\section{Conclusions}
This paper presents RABIT, a versatile conditional image translation \& manipulation framework that adopts a novel bi-level alignment strategy with a ranking and attention scheme (RAS) to align the features between conditional inputs and exemplars efficiently. 
A semantic position encoding mechanism is designed to facilitate semantic-level position information and preserve the texture patterns in the exemplars.
To handle the semantic mismatching between the conditional inputs and warped exemplars, a novel confidence feature injection module is proposed to achieve multi-channel feature fusion based on the matching reliability of warped exemplars.
Quantitative and qualitative experiments over multiple datasets show that RABIT is capable of achieving high-fidelity image translation and manipulation while preserving consistent semantics with the conditional input and faithful styles with the exemplar.

\textbf{Acknowledgement.}
This study is supported under the RIE2020 Industry Alignment Fund – Industry Collaboration Projects (IAF-ICP) Funding Initiative, as well as cash and in-kind contribution from the industry partner(s).


\par\vfill\par

\clearpage
\bibliographystyle{splncs04}
\bibliography{egbib}
\end{document}